\documentclass[11pt]{article}
\usepackage{colacl}
\usepackage[dvips]{graphicx}
\usepackage{xspace}
\usepackage{avm}
\usepackage{examples}
\usepackage{epic}
\usepackage{eepic}
\usepackage{ecltree}

\avmfont{\footnotesize\feat}
\avmvalfont{\footnotesize\fval}
\avmsortfont{\scriptsize\srt}
\avmvskip{.09ex}
\avmhskip{.4em}

\newbox\avmboxa
\newbox\avmboxb
\newbox\avmboxc

\newcommand{\odd}{\sqz{?}}                      
\newcommand{\feat}{\sc}                         
\newcommand{\fval}{\it}                         
\newcommand{\srt}{\it}                          

\newcommand{\hpsg}{\textsc{hpsg}\xspace}

\newcommand{\pref}[1]{(\ref{#1})}               
\newcommand{\qit}[1]{\textsl{``#1''}}           

\title{Selectional Restrictions in HPSG\thanks{
Proceedings of the 18th International Conference on Computational Linguistics
(\textsc{coling}), Saarbr\"ucken, Germany, 31 July -- 4 August 2000, pages
15--20.}}

\author{Ion Androutsopoulos \\
        Software and Knowledge Engineering Laboratory \\
        Institute of Informatics and Telecommunications \\
        National Centre for Scientific \\ Research ``Demokritos'' \\
        153 10 Ag. Paraskevi, Athens, Greece \\
        e-mail: \texttt{ionandr@iit.demokritos.gr}
        \And
        Robert Dale \\
        Language Technology Group \\
        Department of Computing \\
        Macquarie University \\
        Sydney NSW 2109, Australia \\
        e-mail: \texttt{Robert.Dale@mq.edu.au} }
\date{}

\begin{document}
\maketitle

\begin{abstract}
Selectional restrictions are semantic sortal constraints imposed
on the participants of linguistic constructions to capture
contextually-dependent constraints on interpretation. Despite
their limitations, selectional restrictions have proven very
useful in natural language applications, where they have been used
frequently in word sense disambiguation, syntactic disambiguation,
and anaphora resolution. Given their practical value, we explore
two methods to incorporate selectional restrictions in the \hpsg
theory, assuming that the reader is familiar with \hpsg. The first
method employs \hpsg's {\feat background} feature and a
constraint-satisfaction component pipe-lined after the
parser. The second method uses subsorts of referential indices,
and blocks readings that violate selectional restrictions during
parsing. While theoretically less satisfactory, we have found the
second method particularly useful in the development of practical
systems.
\end{abstract}


\section{Introduction} \label{introduction}

The term \emph{selectional restrictions} refers to semantic sortal
constraints imposed on the participants of linguistic constructions.
Selectional restrictions are invoked, for example, to account for the oddity
of \pref{intro:1} and \pref{intro:3} (cf.\ \pref{intro:2} and
\pref{intro:4}).
\begin{examples}
\item \odd Tom ate a keyboard. \label{intro:1}
\item Tom ate a banana. \label{intro:2}
\item \odd Tom repaired the technician. \label{intro:3}
\item Tom repaired the keyboard. \label{intro:4}
\end{examples}
To account for \pref{intro:1} and \pref{intro:2}, one would typically
introduce a constraint requiring the object of \qit{to eat} to denote an
edible entity. The oddity of \pref{intro:1} can then be attributed to a
violation of this constraint, since keyboards are typically not edible.
Similarly, in \pref{intro:3} and \pref{intro:4} one could postulate that
\qit{to repair} can only be used with objects denoting artifacts. This
constraint is violated by \pref{intro:3}, because technicians are typically
persons, and persons are not artifacts.

We note that selectional restrictions attempt to capture
contextually-dependent constraints on interpretation. There is nothing
inherently wrong with \pref{intro:1}, and one can think of special contexts
(e.g.\ where Tom is a circus performer whose act includes gnawing on computer
peripherals) where \pref{intro:1} is felicitous. The oddity of \pref{intro:1}
is due to the fact that in most contexts people do not eat keyboards.
Similarly, \pref{intro:3} is felicitous in a science-fiction context where
the technician is a robot, but not in most usual contexts. Selectional
restrictions are typically used to capture facts about the world which are
generally, but not necessarily, true.

In various forms, selectional restrictions have been used for many
years, and their limitations are well-known \cite{Allen1995}.
For example, they cannot account for metaphoric uses of language
(e.g.\ \pref{intro:4.2}), and they run into problems in negated
sentences (e.g.\ unlike \pref{intro:1}, there is nothing odd about
\pref{intro:4.1}).
\begin{examples}
\item My car drinks gasoline. \label{intro:4.2}
\item Tom cannot eat a keyboard. \label{intro:4.1}
\end{examples}

Despite their limitations, selectional restrictions have proven
very useful in practical applications, and they have been employed
in several large-scale natural language understanding systems
\cite{Martin1983coll} \cite{Alshawi}. Apart from blocking
pragmatically ill-formed sentences like \pref{intro:1} and
\pref{intro:3}, selectional restrictions can also be used in word
sense disambiguation, syntactic disambiguation, and anaphora
resolution. In \pref{intro:5}, for example, the \qit{printer}
refers to a computer peripheral, while in \pref{intro:6} it refers
to a person. The correct sense of \qit{printer} can be chosen in
each case by requiring the object of \qit{to repair} to denote an
artifact, and the subject of \qit{to call} (when referring to a
phone call) to denote a person.
\begin{examples}
\item Tom repaired the printer. \label{intro:5}
\item The printer called this morning. \label{intro:6}
\end{examples}
Similarly, \pref{intro:7} is from a syntactic point of view potentially
ambiguous: the relative clause may refer either to the departments or the
employees. The correct reading can be chosen by specifying that the subject
of \qit{retire} (the relativised nominal in this case) must denote a person.
\begin{examples}
\item List the employees of the overseas departments that will
retire next year. \label{intro:7}
\end{examples}

Given the value of selectional restrictions in practical
applications, we explore how they can be utilised in the \hpsg
theory \cite{Pollard2}, assuming that the reader is familiar with
\hpsg. Our proposals are based on experience obtained from using
\hpsg in a natural language database interface
\cite{Androutsopoulos1998a} and a dialogue system for a mobile
robot. To the best of our knowledge, selectional restrictions have
not been explored so far in the context of \hpsg.

We note that, although they often exploit similar techniques
(e.g.\ semantic sort hierarchies), selectional restrictions
costitute a different topic from \emph{linking theories}
\cite{Davis1996}. Roughly speaking, linking theories explore the
relation between thematic roles (e.g.\ agent, patient) and
grammatical functions (e.g.\ subject, complement), while
selectional restrictions attempt to account for the types of world
entities that can fill the thematic roles.

We discuss in sections 2 and 3 two ways that we have considered
to incorporate selectional restrictions into \hpsg. Section
\ref{conclusions} concludes by comparing briefly the two approaches.


\section{Background restrictions} \label{background_restrictions}

The first way to accommodate selectional restrictions in \hpsg
uses the {\feat context$\mid$background} (abbreviated here as
{\feat cx$\mid$bg}) feature, which Pollard and Sag \cite{Pollard2}
reserve for ``felicity conditions on the utterance context'',
``presuppositions or conventional implicatures'', and
``appropriateness conditions'' ({\it op cit} pp.~27, 332). To
express selectional restrictions, we add qfpsoas (quantifier-free
parameterised states of affairs) with a single semantic role
(slot) in {\feat cx$\mid$bg}.\footnote{To save space, we use
qfpsoas wherever Pollard and Sag use quantified psoas. We also
ignore tense and aspect here. Consult \cite{Androutsopoulos1998a}
for the treatment of tense and aspect in our \hpsg-based database
interface.} For example, apart from the {\srt eat}\/ qfpsoa  in
its {\feat nucleus} ({\feat nuc}), the lexical sign for \qit{ate}
(shown in \pref{background:1}) would introduce an {\srt edible}\/
qfpsoa in {\feat bg}, requiring \avmbox{2} (the entity denoted by
the object of \qit{ate}) to be edible.
\begin{examples}
\item \avmoptions{active}
\begin{avm}
[\avmspan{phon \; \<\fval ate\>} \\
 ss|loc & [cat & [head  & verb \\
                  subj   & \< \feat np$_{@1}$ \>  \\
                  comps  & \< \feat np$_{@2}$ \> ]\\
           cont|nuc & \osort{eat}{
                      [eater & @1 \\
                       eaten & @2]} \\
           cx|bg & \{\sort{edible}{
                       [inst & @2]} \}]]
\end{avm}
\label{background:1}
\end{examples}

In the case of lexical signs for proper names (e.g.\
\pref{background:2}), the treatment of Pollard and Sag inserts a
{\srt naming}\/ ({\srt namg}\/) qfpsoa in {\feat bg}, which
requires the {\feat bearer} ({\feat brer}) to by identifiable in
the context by means of the proper name. \pref{background:2} also
requires the bearer to be a man.

\begin{examples}
\item \avmoptions{active}
\begin{avm}
[\avmspan{phon \; \<\fval Tom\>} \\
 ss|loc & [cat & [head  & noun] \\
           cont & [index  & @1\\
                   restr & \{ \}] \\
           cx|bg & \{\sort{namg}{
                     [brer & @1 \\
                      name & Tom]}, \\
                     \sort{man}{
                     [inst & @1]} \}]]
\end{avm}
\label{background:2}
\end{examples}

The \hpsg principles that control the propagation of the {\feat
bg} feature are not fully developed. For our purposes, however,
the simplistic \emph{principle of contextual consistency} of
Pollard and Sag will suffice. This principle causes the {\feat bg}
value of each phrase to be the union of the {\feat bg} values of
its daughters. Assuming that the lexical sign of \qit{keyboard} is
\pref{background:2.1}, \pref{background:1}--\pref{background:2.1}
cause \pref{intro:1} to receive \pref{new:1}, that requires
\avmbox{2} to denote an edible keyboard.

\begin{examples}
\item \avmoptions{active}
\begin{avm}
[\avmspan{phon \; \<\fval keyboard\>} \\
 ss|loc & [cat & [head  & noun] \\
           cont & [index  & @2\\
                   restr & \{ \sort{keybd}{
                              [inst & @2]} \}] \\
           \avmspan{\feat cx|bg \; \{ \}}]]
\end{avm}
\label{background:2.1}

\item \avmoptions{active}
\begin{avm}
[phon   & \<\fval Tom, ate, a, keyboard\> \\
 ss|loc & [cat   & [head   & verb \\
                    subj   & \< \> \\
                    comps  & \< \>] \\
           cont  & [quants & \< \sort{keybd}{
                                [inst & @2]} \> \\
                    nuc    & \osort{eat}{
                             [eater & @1 \\
                              eaten & @2]}] \\
           cx|bg & \{\sort{namg}{
                     [brer & @1 \\
                      name & Tom]}, \\
                     \sort{man}{
                     [inst & @1]}, \\
                     \sort{edible}{
                     [inst & @2]} \}]]
\end{avm}
\label{new:1}
\end{examples}

According to \pref{new:1}, to accept \pref{intro:1}, one has to
place it in a special context where edible keyboards exist (e.g.\
\pref{intro:1} is felicitous if it refers to a miniature chocolate
keyboard). Such contexts, however, are rare, and hence
\pref{intro:1} sounds generally odd. Alternatively, one has to
relax the {\feat bg} constraint that the keyboard must be edible.
We assume that special contexts allow particular {\feat bg}
constraints to be relaxed (this is how we would account for the
use of \pref{intro:1} in a circus context), but we do not have any
formal mechanism to specify exactly when {\feat bg} constraints
can be relaxed.

Similar comments apply to \pref{intro:3}. Assuming that the sign
of \qit{repaired} is \pref{background:6}, and that the sign of
\qit{technician} is similar to \pref{background:2.1} except that
it introduces a {\srt technician}\/ index, \pref{intro:3} receives
a sign that requires the repairer to be a technician who is an
artifact. Technicians, however, are generally not artifacts, which
accounts for the oddity of \pref{intro:3}.
\begin{examples}
\item \avmoptions{active}
\begin{avm}
[\avmspan{phon \; \<\fval repaired\>} \\
 ss|loc & [cat & [head  & verb \\
                  subj   & \< \feat np$_{@1}$ \>  \\
                  comps  & \< \feat np$_{@2}$ \> ]\\
           cont|nuc & \sort{repair}{
                      [repairer & @1 \\
                       repaired & @2]} \\
           cx|bg & \{\sort{artifact}{
                       [inst & @2]} \}]]
\end{avm}
\label{background:6}
\end{examples}

\begin{figure}
\hrule
\medskip
\begin{center}
\includegraphics[scale=.5]{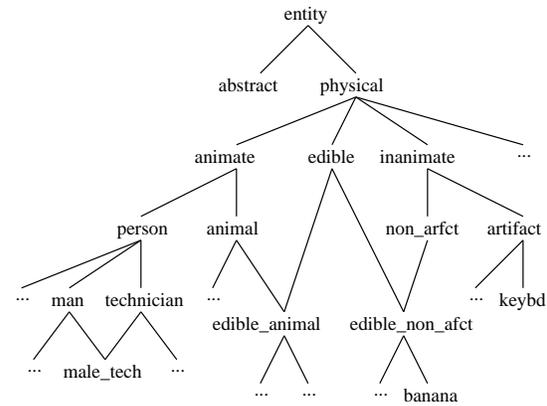}
\caption{A simplistic semantic hierarchy}
\label{hierarchy_fig}
\end{center}
\hrule
\end{figure}

Let us now consider how a computer system could account for
\pref{intro:1}--\pref{intro:4}. For example, how would the system
figure out from \pref{new:1} that \pref{intro:1} is pragmatically
odd? Among other things, it would need to know that keyboards are
not edible. Similarly, in \pref{intro:2} it would need to know
that bananas \emph{are} edible, and in
\pref{intro:3}--\pref{intro:4} it would need to be aware that
technicians are not artifacts, while keyboards are. Systems that
employ selectional restrictions usually encode knowledge of this
kind in the form of sort hierarchies of world entities. A
simplistic example of such a hierarchy is depicted in figure
\ref{hierarchy_fig}. The hierarchy of figure \ref{hierarchy_fig}
shows that all men and technicians are persons, all persons are
animate entities, all animate entities are physical objects, and
so on. Some (but not all) persons are both technicians and men at
the same time; these persons are members of the
$\mathit{male\_tech}$ sort. Similarly, all bananas are edible and
not artifacts. No person is edible, because the sorts
$\mathit{person}$ and $\mathit{edible}$ have no common subsorts.

It is, of course, extremely difficult to construct hierarchies
that include \emph{all} the sorts of world entities. In natural
language systems that target specific and restricted domains,
however, constructing such hierarchies is feasible, because the
relevant entity sorts and the possible hierarchical relations
between them are limited. In natural language database interfaces,
for example, the relevant entity sorts and the relations between
them are often identified during the desing of the database, in
the form of entity-relationship diagrams. We also note that
large-scale semantic sort hierarchies are already in use in
artificial intelligence and natural language generation projects
(for example, Cyc \cite{Lenat1995} and \textsc{kpml}'s Upper Model
\cite{Bateman1997}), and that the techniques that we discuss in
this paper are in principle compatible with these hierarchies.

To decide whether or not a sentence violates any selectional restrictions, we
collect from the {\feat cont} and {\feat bg} features of its sign
(\pref{new:1} in the case of \pref{intro:1})
all the single-role qfpsoas for which there is a sort in the hierarchy with
the same name. (This rules out single-slot qfpsoas introduced by the {\feat
cont}s of intransitive verbs.) The decision can then be seen as a
constraint-satisfaction problem, with the collected qfpsoas acting as
constraints. \pref{background:7} shows the constraints for \pref{intro:1},
rewritten in a form closer to predicate logic. \hpsg indices (the boxed
numbers) are used as variables.
\begin{examples}
\item $\mathit{keybd}(\avmbox{2}) \land \mathit{man}(\avmbox{1}) \land
\mathit{edible}(\avmbox{2})$ \label{background:7}
\end{examples}

Given two contstraints $c_1, c_2$ on the same variable, $c_1$
\emph{subsumes} $c_2$ if the corresponding hierarchy sort of $c_1$
is an ancestor of that of $c_2$ or if $c_1 = c_2$. $c_1$ and $c_2$
can be replaced by a new single constraint $c$, if $c_1$ and $c_2$
subsume $c$, and there is no other constraint $c'$ which is
subsumed by $c_1, c_2$ and subsumes $c$. $c$ and $c'$ must be
constraints on the same variable as $c_1, c_2$, and must each correspond
to a sort of the hierarchy. If the constraints of a sentence can be
turned in this way into a form where there is only one constraint
for each variable, then (and only then) the sentence violates no
selectional restrictions.

In \pref{background:7}, $\mathit{keybd}(\avmbox{2})$ and
$\mathit{edible}(\avmbox{2})$ cannot be replaced by a single
constraint, because {\srt keybd}\/ and {\srt edible}\/ have no
common subsorts. Hence, a selectional restriction is violated,
which accounts for the oddity of \pref{intro:1}. In contrast, in
\pref{intro:2} the constraints would be as in \pref{background:8}.
\begin{examples}
\item $\mathit{banana}(\avmbox{2}) \land \mathit{man}(\avmbox{1}) \land
\mathit{edible}(\avmbox{2})$ \label{background:8}
\end{examples}
$\mathit{banana}(\avmbox{2})$ and $\mathit{edible}(\avmbox{2})$
can now be replaced by $\mathit{banana}(\avmbox{2})$, because both
subsume $\mathit{banana}(\avmbox{2})$, and no other constraint
subsumed by both $\mathit{banana}(\avmbox{2})$ and
$\mathit{edible}(\avmbox{2})$ subsumes
$\mathit{banana}(\avmbox{2})$. This leads to \pref{background:9}
and the conclusion that \pref{intro:2} does not violate any
selectional restrictions.
\begin{examples}
\item $\mathit{banana}(\avmbox{2}) \land \mathit{man}(\avmbox{1})$
\label{background:9}
\end{examples}

This constraint-satisfaction reasoning, however, requires a
separate inferencing component that would be pipe-lined after the
parser to rule out signs corresponding to sentences (or readings)
that violate selectional restrictions. In the next section, we
discuss an alternative approach that allows hierarchies of world
entities to be represented using the existing \hpsg framework, and
to be exploited during parsing without an additional
inferencing component.


\section{Index subsorts} \label{index_subsorts}

\hpsg has already a hierarchy of feature structure sorts \cite{Pollard2}.
This hierarchy can be augmented to include a new part that encodes
information about the types of entities that exist in the world. This can be
achieved by partitioning the {\srt ref}\/ \hpsg sort (currently, a leaf node
of the hierarchy of feature structures that contains all indices that refer
to world entities) into subsorts that correspond to entity types. To encode
the information of figure \ref{hierarchy_fig}, {\srt ref}\/ would have the
subsorts {\srt abstract}\/ and {\srt physical}, {\srt physical}\/ would have
the subsorts {\srt animate}, {\srt edible}, {\srt inanimate}, and so on. That
is, referential indices are partitioned into sorts, and the indices of each
sort can only be anchored to world entities of the corresponding type (e.g.\
{\srt keybd}\/ indices can only be anchored to keyboards).

With this arrangement, the lexical sign for \qit{ate} becomes
\pref{subsorts:1}. The {\feat bg} {\srt edible}\/ restriction of
\pref{background:1} has been replaced by the restriction that the
index of the object must be of sort {\srt edible}.
\begin{examples}
\item \avmoptions{active}
\begin{avm}
[\avmspan{phon \; \<\fval ate\>} \\
 ss|loc & [cat & [head  & verb \\
                  subj   & \< \feat np$_{@1}$ \>  \\
                  comps  & \< \feat np$_{@2}$ \> ]\\
           cont|nuc & \osort{eat}{
                      [eater & @1 \\
                       eaten & @2edible]}]]
\end{avm}
\label{subsorts:1}
\end{examples}
Similarly, the sign for \qit{Tom} becomes \pref{subsorts:2} (cf.\
\pref{background:2}), and the sign for \qit{keyboard} introduces
an index of sort {\srt keybd}\/ as shown in \pref{new:3} (cf.\
\pref{background:2.1}).
\begin{examples}
\item \avmoptions{active}
\begin{avm}
[\avmspan{phon \; \<\fval Tom\>} \\
 ss|loc & [cat & [head  & noun] \\
           cont & [index  & @1man\\
                   restr & \{ \}] \\
           cx|bg & \{\sort{namg}{
                     [brer & @1 \\
                      name & Tom]}\}]]
\end{avm}
\label{subsorts:2}

\item \avmoptions{active}
\begin{avm}
[\avmspan{phon \; \<\fval keyboard\>} \\
 ss|loc & [cat & [head  & noun] \\
           cont & [index  & @2keybd\\
                   restr & \{ \}] \\
           \avmspan{\feat cx|bg \; \{ \}}]]
\end{avm}
\label{new:3}
\end{examples}

Unification of indices proceeds in the same manner as unification
of all other typed feature structures \cite{Carpenter1992}. The
parsing of \pref{intro:1} now fails, because it attempts to unify
an index of sort {\srt edible}\/ (introduced by \pref{subsorts:1})
with an index of sort {\srt keybd}\/ (introduced by \pref{new:3}),
and no \hpsg sort is subsumed by both. In contrast, the parsing of
\pref{intro:2} would succeed, because the sign of \qit{banana}
would introduce an index of sort {\srt banana}, which is a subsort
of {\srt edible} (figure \ref{hierarchy_fig}); hence the two
indices can be unified. \pref{intro:3} and \pref{intro:4} would be
processed similarly.

In \pref{intro:5} and \pref{intro:6}, there would be two lexical
signs for \qit{printer}: one introducing an index of sort
$\mathit{printer\_person}$, and one inroducing an index of sort
$\mathit{printer\_peripheral}$. ($\mathit{printer\_person}$ and
$\mathit{printer\_peripheral}$ would be daughters of
$\mathit{person}$ and $\mathit{artifact}$ respectively in figure
\ref{hierarchy_fig}.) The sign for \qit{repaired}, would require
the index of its object to be of sort $\mathit{artifact}$, and the
sign of \qit{called} would require its subject index to be of sort
$\mathit{person}$. This correctly admits only the reading where
the repaired entity is a computer peripheral, and the caller is a
person. Similar mechanisms can be used to determine the correct
reading of \pref{intro:7}.

\begin{figure}[tb]
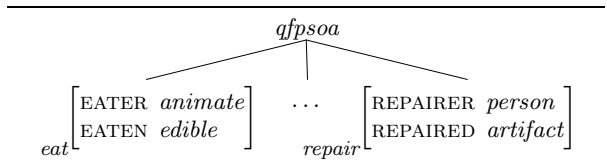

\avmoptions{}
\setlength{\GapWidth}{5mm}
\hrule
\begin{center}
{\footnotesize
\begin{bundle}{{\srt qfpsoa}}
\chunk{\begin{avm}
       \sort{eat}{
       \[eater & animate \\
         eaten & edible\]}
       \end{avm}}
\chunk{\dots}
\chunk{\begin{avm}
       \osort{repair}{
       \[repairer & person \\
         repaired & artifact\]}
       \end{avm}}
\end{bundle}
}
\caption{Declarations of qfpsoas}
\label{qfpsoas_fig}
\end{center}
\hrule
\end{figure}

With the approach of this section, it is also possible to specify
selectional restrictions in the declarations of qfpsoas in the
\hpsg hierarchy of feature structures, as shown in figure
\ref{qfpsoas_fig}, rather than in the lexicon.\footnote{Additional
layers can be included between {\srt qfpsoa}\/ and the leaf sorts,
as sketched in section 8.5 of \cite{Pollard2}, to group together
qfpsoas with common semantic roles.} When the same qfpsoa is used
in several lexical signs, this saves having to repeat the same
selectional restrictions in each one of the lexical signs. For
example, the verbs \qit{repair} and \qit{fix} may both introduce a
{\srt repair}\/ qfpsoa. The restriction that the repaired entity
must be an artifact can be specified once in the declaration of
{\srt repair}\/ in the hierarchy of feature structures, rather
than twice in the lexical signs of \qit{repair} and \qit{fix}.


\section{Conclusions} \label{conclusions}

We have presented two methods to incorporate selectional
restrictions in \hpsg: (i) expressing selectional restrictions as
{\feat background} constraints, and (ii) employing subsorts of
referential indices. The first method has the advantage that it
requires no modification of the current \hpsg feature structures.
It also maintains Pollard and Sag's distinction between
``literal'' and ``non-literal'' meaning (expressed by {\feat cont}
and {\feat background} respectively), a distinction which is
blurred in the second approach (e.g.\ nothing in \pref{subsorts:1}
shows that requiring the object to denote an edible entity is part
of the non-literal meaning; cf.\ \pref{background:1}). Unlike the
first method, however, the second approach requires no additional
inferencing component for determining when selectional
restrictions have been violated. With sentences that contain
several potentially ambiguous words or phrases, the second
approach is also more efficient, as it blocks signs that violate
selectional restrictions during parsing. In the first approach,
these signs remain undetected during parsing, and they may have a
multiplicative effect, leading to a large number of parses, which
then have to be checked individually by the inferencing component.
We have found the second approach particularly useful in the
development of practical systems.

There is a deeper question here about the proper place to maintain
the kind of information encoded in selectional restrictions. The
applicability of selectional restrictions is always
context-dependent; and for any selectional restriction, we can
almost always find a context where it does not hold. Our second
method above effectively admits that we cannot develop a general
purpose solution to the problem of meaning interpretation, and
that we have to accept that our systems always operate in specific
contexts. By committing to a particular context of interpretation,
we `compile into' what was traditionally thought of as literal
meaning a set of contextually-determined constraints, and thus
enable these constraints to assist in the \hpsg language analysis
without requiring an additional reasoning component. We take the
view here that this latter approach is very appropriate in the
construction of real applications which are, and are likely to be
for the foreseeable future, restricted to operating in limited
domains.


\bibliographystyle{acl}
\bibliography{biblio}

\begin{thebibliography}{}

\bibitem[\protect\citename{Allen}1995]{Allen1995}
J.F. Allen.
\newblock 1995.
\newblock {\em {Natural Language Understanding}}.
\newblock Benjamin/Cummings.

\bibitem[\protect\citename{Alshawi}1992]{Alshawi}
H.~Alshawi, editor.
\newblock 1992.
\newblock {\em {The Core Language Engine}}. MIT Press.

\bibitem[\protect\citename{Androutsopoulos \bgroup et al.\egroup
  }1998]{Androutsopoulos1998a}
I.~Androutsopoulos, G.D. Ritchie, and P.~Thanisch.
\newblock 1998.
\newblock {Time, Tense and Aspect in Natural Language Database Interfaces}.
\newblock {\em Natural Language Engineering}, 4(3):229--276.

\bibitem[\protect\citename{Bateman}1997]{Bateman1997}
J.A. Bateman.
\newblock 1997.
\newblock {Enabling Technology for Multilingual Natural Language Generation:
  the KPML Development Environment}.
\newblock {\em {Natural Language Engineering}}, 3(1):15--55.

\bibitem[\protect\citename{Carpenter}1992]{Carpenter1992}
B.~Carpenter.
\newblock 1992.
\newblock {\em {The Logic of Typed Feature Structures}}.
\newblock Number~32 in Cambridge Tracts in Theoretical Computer Science.
  Cambridge University Press.

\bibitem[\protect\citename{Davis}1996]{Davis1996}
T.~Davis.
\newblock 1996.
\newblock {\em {Lexical Semantics and Linking in the Hierarchical Lexicon}}.
\newblock {Ph.D.} thesis, Stanford University.

\bibitem[\protect\citename{Lenat}1995]{Lenat1995}
D.B. Lenat.
\newblock 1995.
\newblock {CYC: A Large-Scale Investment in Knowledge Infrastructure}.
\newblock {\em Communications of ACM}, 38(11):33--38.

\bibitem[\protect\citename{Martin \bgroup et al.\egroup }1986]{Martin1983coll}
P.~Martin, D.~Appelt, and F.~Pereira.
\newblock 1986.
\newblock {Transportability and Generality in a Natural-Language Interface
  System}.
\newblock In B.~Grosz, K.~Sparck~Jones, and B.~Webber, editors, {\em {Readings
  in Natural Language Processing}}, pages 585--593. Morgan Kaufmann.

\bibitem[\protect\citename{Pollard and Sag}1994]{Pollard2}
C.~Pollard and I.A. Sag.
\newblock 1994.
\newblock {\em {Head-Driven Phrase Structure Grammar}}.
\newblock University of Chicago Press and Center for the Study of Language and
  Information, Stanford.

\end{thebibliography}

\end{document}